\documentclass{pf2010}

\usepackage{pstricks}
\usepackage{verbatim}
\usepackage{subfigure}
\usepackage{algorithmic}
\usepackage{algorithm}

\usepackage[french]{babel}
\usepackage[latin1]{inputenc}
\usepackage[T1]{fontenc}
\usepackage{amssymb}
\usepackage{amsmath}

\usepackage{natbib}

\shorttitle{Filtrage vaste marge}

\shortouvrage{CAp 2010}

\newcommand{\swp}[2]{#2}

\newcommand{\altern}[2]{{\color{blue} #2}}

\def\X{\ensuremath{X}}

\def\F{\ensuremath{F}}
\def\w{\ensuremath{\mathbf{d}}}

\def\Xf{\ensuremath{\widetilde{X}}}

\def\Kf{\ensuremath{\widetilde{K}}}

\def\Xfte{\ensuremath{\widetilde{Xte}}}
\def\y{\ensuremath{\mathbf{y}}}

\def\dbR{{\mathrm{I\hskip-2.2pt R}}}
\def\dbR{\mathbb{R}}

\def\Eq{Equation }

\title{Filtrage vaste marge pour l'étiquetage séquentiel à noyaux de signaux}

\author{R\'emi Flamary, Benjamin Labb\'e, Alain Rakotomamonjy
\institute{LITIS EA 4108, INSA-Universit\'e de Rouen\\76801
  Saint-\'Etienne du Rouvray Cedex,
  FRANCE
}
}

\begin{document}
\maketitle

\begin{abstract}
Ce papier traite de l'\'etiquetage s\'equentiel de signaux,
c'est-\`a-dire de discrimination pour des \'echantillons temporels. 
Dans ce contexte, nous proposons
une méthode d'apprentissage pour un filtrage vaste-marge 
s\'eparant au mieux les classes. Nous apprenons ainsi de
manière jointe un SVM sur des \'echantillons et un filtrage temporel de
ces \'echantillons. Cette m\'ethode permet l'\'etiquetage en ligne
d'\'echantillons temporels. Un d\'ecodage de s\'equence hors ligne 
optimal utilisant l'algorithme de Viterbi est \'egalement propos\'e.
Nous introduisons diff\'erents termes de régularisation,
permettant de pondérer ou de
s\'electionner les canaux automatiquement au sens du crit\`ere vaste-marge. 
Finalement, notre approche est test\'ee sur un exemple jouet de signaux non-lin\'eaires
ainsi que sur des donn\'ees réelles d'Interface Cerveau-Machine.
Ces expériences montrent
l'int\'er\^et de l'apprentissage supervis\'e d'un filtrage temporel pour
l'\'etiquetage de s\'equence.

\motscles{SVM, \'Etiquetage séquentiel, Filtrage}
\end{abstract}

\section{Introduction}
\label{sec:introduction}

Signal sequence labeling is a classical machine learning problem that
typically arises in Automatic Speech Recognition (ASR) or Brain
Computer Interfaces (BCI). The idea is to assign a label for every
sample of a signal while taking into account the sequentiality of the
samples.  
For instance, in speaker diarization, the aim is to recognize which
speaker is talking along time. 
Another example is the recognition of mental states from
Electro-Encephalographic (EEG) signals. This mental states are then
mapped into commands for a computer (virtual keyboard, mouse) or a
mobile robot, hence the need for sample labeling \cite{bcicometitioniii,millan04}.


One widely used approach for performing sequence labeling is Hidden
Markov Models (HMMs), cf.  \citep{infmakov}. HMMs are probabilistic
models that may be used for sequence decoding of discrete states
observations.  In the case of continuous observations such as signal
samples or vectorial features extracted from the signal, Continuous
Density HMMs \altern{(CDHMM)}{} are
considered.
When using HMM for sequence decoding, one needs to have the
conditional probability of the observations per hidden states
(classes), which is usually obtained through Gaussian Mixtures
(GM) \citep{infmakov}. But this kind of model performs poorly in high
dimensional spaces in terms of discrimination, and recent works have shown that
the decoding accuracy may be improved by using discriminative
models~\citep{\swp{altun2003hidden,}{}sloin2008support}. One simple approach
for using discriminative classifiers in the HMM framework has been
proposed by \citet{ganapathiraju2004applications}. It consists in
learning SVM classifiers known for their better robustness in high
dimension and to transform their outputs to probabilities using
Platt's method \citep{lin2007note}, leading to
better performances after Viterbi decoding.
However, this approach supposes that the
complete sequence of observation is available, which corresponds to an
offline decoding. In the case of BCI application, a real time decision
is often needed \citep{bcicometitioniii,millan04}, which restricts the
use of the Viterbi decoding.

Another limit of HMM is that they cannot take into account 
a time-lag between the labels and the
discriminative features. 
Indeed, in this case some of the learning observations are
mislabeled, leading to a biased density estimation per class.
This is a problem in BCI applications where the interesting
information are not always synchronized with the labels. 
For instance, \citet{Pistohl2008} showed the need of applying delays
to the signal, since the neuronal activity precedes the actual
movement. 
Note that they selected the delay through validation.
Another illustration of the need of time-lag automated handling 
is the following. Suppose we want to interact
with a computer using multi-modal acquisitions (EEG,EMG,\dots).
Then, since each modality has its own time-lag with respect to neural 
activity as shown by \citet{salenius1996human}, it may be
difficult to manually synchronize all modalities and better
adaptation can be obtained by learning the ``best'' time-lag
to apply to each modality channel.



Furthermore, instead of using a fixed filter as a preprocessing stage
for signal denoising, learning the filter may help in adapting
to noise characteristics of each channel in addition to the
time-lag adjustment.
In such a context, \citet{flamaryicassp210} proposed a method to learn a large margin 
filtering for linear SVM classification of samples (FilterSVM). They learn a
Finite Impulse Response (FIR) filter for each channel of the signal jointly with a linear
classifier. 
Such an approach has the flavor
of the Common Sparse-Spatio-Spectral Pattern (CSSSP) of
\citet{dornhege2006optimizing} 
as it corresponds to a filter which helps
in discriminating classes. 
However, CSSSP is a supervised feature extraction method based on
time-windows, whereas FilterSVM is a sequential sample classification
method.
Moreover, the unique temporal filter provided by CSSSP cannot adapt to
different channel properties, at the contrary of FilterSVM that learns one filter per channel.


In this paper, we extend the work of
\citet{flamaryicassp210} to the non-linear case.
We propose algorithms that may be used to obtain
large margin filtering in non-linear problems. 
Moreover, we study and discuss the effect of different regularizers for
the filtering matrix. 
Finally, in the experimental section we test our
approach on a toy example for online and offline decision (with a
Viterbi decoding) and investigate the parameters sensitivity of
our method. 
We also benchmark our approach in a online
sequence labeling situation by means of a BCI problem. 



\section{Sample Labeling}
\label{sec:method}

First we define the problem of sample labeling and the filtering of a
multi-dimensionnal signal. Then we define the SVM classifier for
filtered samples.
\subsection{Problem definition}


We want to obtain a
sequence of labels from a  multi-channel signal
or from multi-channel features extracted from
that signal. 
We suppose that the training samples  are gathered in
a matrix 
 $\X\in\dbR^{n\times d}$ containing $d$ channels and $n$ samples.  
$\X_{i,v}$~is the value of channel~$v$ for
the $i^{th}$ sample. The vector $\y\in\{-1,1\}^n$ contains the class of each sample.



In order to reduce noise in the samples
or variability in the features, a usual approach is to filter $\X$
before the classifier learning stage. In literature,
all channels are usually filtered with the same filter
(\citet{Pistohl2008} used a Savisky-Golay filter) although there is no
reason
for a single filter to be optimal for all  channels. Let us
define the filter applied to $\X$ by the matrix $\F\in \dbR^{f\times
  d}$. Each column of $\F$ is a filter for the corresponding channel
in $\X$ and $f$ is the size of the 
filters. 

We define the filtered data matrix $\Xf$ by:
\begin{equation}
  \label{eq:1}
  \Xf_{i,v}=\sum_{u=1}^f ~ \F_{u,v} ~ \X_{i+1-u+n_0,v}\swp{=\X_{i,v}\ast\F_{.,v}}{}
\end{equation}
where the sum is a unidimensional convolution \swp{($\ast$)}{} of each channel by the filter in
the appropriate column of $F$. $n_0$ is the delay of the filter, for
instance $n_0=0$ corresponds to a causal filter and  $n_0=f/2$
corresponds to a filter centered on the current sample.


\subsection{SVM for filtered samples}
\label{sec:averaged SVM}

A good way of improving the classification rate is to filter the channels in $\X$ in
order to reduce the impact of the noise. The simplest filter in the 
case of high frequency noise is the average filter 
defined by $\F_{v,u}=1/f, \forall i \in \{1,\dots,f\}$ and $j \in
\{1,\dots,d\}$. $n_0$ is selected depending on the problem at hand,
$n_0$=0 for a causal filtering of $n_0>0$ for a non-causal filtering.
In the following, using an average filter as preprocessing on the
signal and an SVM classifier will be called Avg-SVM.

Once the filtering is chosen we can learn an SVM sample classifier on
the filtered samples by solving the problem:
\begin{equation}
  \label{eq:AveragedSVM}
  \min_{g} \quad\frac{1}{2} ||g||^2+\frac{C}{n}\sum_{i=1}^n H(\y_i,\Xf_{i,.},g)
\end{equation}
where $C$  is the regularization parameter, $g(\cdot)$ is the decision function and
$H(y,x,g)=\max(0,1-y\cdot g(x))$ is the hinge loss. In practice for
non-linear case, one solve the dual form of this problem wrt. $g$:
\begin{eqnarray}
  \label{eq:dualsvm}
  \max_\alpha J_{SVM}(\alpha,F)=\swp{}{\max_\alpha}-\sum_{i,j}^{n,n}
  \y_i\y_j\alpha_i\alpha_j\Kf_{i,j}+\sum_i^N\alpha_i\\
s.t.\quad \frac{C}{n}\geq\alpha_i\geq0 \quad \forall i \quad and \quad \sum_i^N\alpha_i\y_i=0 \nonumber
\end{eqnarray}
where $\forall i\in [1,n] , \alpha_i\in \dbR$ are the dual variables and $\Kf$ is the kernel
matrix for filtered samples in the
gaussian case. When $\sigma_k$ is the kernel bandwidth, $\Kf$ is
defined by:
\swp{
\begin{align}
  \label{eq:kernelmatrix}
  \Kf_{i,j}&=&k(\Xf_{i,.},\Xf_{j,.})=\exp\left(-\frac{||\Xf_{i,.}-\Xf_{j,.}||^2}{2\sigma_k^2}\right)\\
 & =&\exp\left(-\frac{\sum_v^d||(\X_{i,v}-\X_{j,m})\ast\F_{.,v}||^2}{2\sigma_k^2}\right)
\end{align}}{
\begin{equation}
  \label{eq:kernelmatrix}
  \Kf_{i,j}=k(\Xf_{i,.},\Xf_{j,.})=\exp\left(-\frac{||\Xf_{i,.}-\Xf_{j,.}||^2}{2\sigma_k^2}\right)
\end{equation}}
Note that for any FIR filter, the resulting $\Kf$ matrix is always positive
definite if $k(\cdot,\cdot)$ is definite positive.
Indeed, suppose $k(\cdot,\cdot)$ a kernel from $\mathcal{X}^2$ to $\dbR$ and
$\phi$ a mapping from any $\mathcal{X}'$ to $\mathcal{X}$, then
$k'(\cdot,\cdot)=k(\phi(\cdot),\phi(\cdot))$ is a positive definite
kernel \swp{\cite{shawe2004kernel}}{}. Here, our filter is a linear
combination of $\dbR^d$ elements, which is still in $\dbR^d$. 

Once the classifier is learned, the decision function for a new filtered
signal $\Xfte$ at sample $i$ is:
\begin{equation}
  \label{eq:5}
  g(i,\Xfte)=\sum_{j=1}^n\alpha_j y_j k(\Xfte_{i,.},\Xf_{j,.})
\end{equation}
We show in the experiment section that this approach leads to
improvement over the usual non-filtered approach. But the methods
rely on the choice of a filter depending on prior information or user
knowledge. And there is no evidence that the user-selected filter will be
optimal in any sense for a given classification task.

\section{Large Margin Filtering for non-linear problems (KF-SVM)}
\label{sec:large-marg-filt}
\swp{
Here, we present the optimization problem we want to solve
and the proposed algorithms . Then we discuss the use of
different regularizers and the related works.

\subsection{Optimization problem}
\label{sec:optimization-problem}}{}

We propose in this section to jointly learn the filtering matrix $\F$
and the classifier, this method will be named KF-SVM in the
following. 
It leads to a filter maximizing the
margin between the classes in the feature space.
The problem we want to solve is:
\begin{equation}
  \label{eq:filtersvmcost}
  \min_{g,\F}\quad \frac{1}{2} ||g||^2+\frac{C}{n}\sum_{i=1}^n H(\y_i,\Xf_{i,.},g)+\lambda\Omega (\F)
\end{equation}
with $\lambda$ a regularization parameter and $\Omega(\cdot)$ a differentiable
regularization function of $\F$.
We can recognize in the left part of
\Eq \eqref{eq:filtersvmcost} a SVM problem for filtered samples $\Xf$
but with $\F$ as a variable.
This objective function is
non-convex. 
However, for a fixed $\F$, the optimization problem wrt. $g(\cdot)$ is
convex and boils down to a SVM problem.
So we propose to solve \Eq \eqref{eq:filtersvmcost} by a
coordinate-wise approach:
\begin{equation}
 \label{eq:filtersvmcost2}
  \min_{\F} J(F)=\min_{\F}J'(\F)+\lambda\Omega (\F)
\end{equation}
with:
\begin{align}
J'(\F)&=\min_g \quad \frac{1}{2}
||g||^2+\frac{C}{n}\sum_{i=1}^n H(\y_i,\Xf_{i,.},g)\label{eq:jp_primal}\\
&=\max_{C/n\geq\alpha\geq0,\sum_i \alpha_i\y_i=0} J_{SVM}(\alpha,\F)\label{eq:jp_dual}
\end{align}
where $J_{SVM}$ is defined in \Eq \eqref{eq:dualsvm} and $g(\cdot)$ is  defined in
\Eq \eqref{eq:5}.
Due to the strong duality of the SVM problem, $J'(\cdot)$
can be expressed in his primal or dual form (see
\eqref{eq:jp_primal} and \eqref{eq:jp_dual}).
The objective function $J$ defined in \Eq \eqref{eq:filtersvmcost2} is
non-convex. 
But according to \citet{bonnans_pertubation} for a given $F^*$,
$J'(\cdot)$ is differentiable wrt. $F$. At 
the point $F^*$, the gradient of $J(\cdot)$ can be
computed. Finally we
can solve the problem in \Eq \eqref{eq:filtersvmcost2} by doing a gradient
descent on $J(F)$ along $\F$. 

Note that due to  the 
 non-convexity of the objective functions, problems
 (\ref{eq:filtersvmcost}) and \eqref{eq:filtersvmcost2} are not
 strictly equivalent. But its advantageous to solve
 \eqref{eq:filtersvmcost2} because it can be solved using SVM solvers
 and our method would benefit from any improvement in this domain.

\subsection{KF-SVM Solver and complexity}
\label{sec:kf-svm-solver}

For solving the optimization
problem, we propose a conjugate gradient (CG) descent algorithm along $\F$ with a line
search method for finding the optimal step.
The method is detailed in Algorithm~\ref{algorithm1}, where
$\beta$ is the CG update parameter and $D_\F^i$ the descent
direction for the $i$th iteration. 
For the experimental results we
used the $\beta$ proposed by Fletcher and Reeves, see
\citep{hager2006survey} for more information.
The iterations in the algorithm may be stopped by two stopping criteria: a
 threshold on the relative variation of $J(F)$ or on the
 norm of the variation of $\F$. 
 \begin{algorithm}[htb]
 \caption{ KF-SVM solver \label{algorithm1}}
 \begin{algorithmic}
 \STATE Set $\F_{u,v}=1/f$ for $v=1\cdots d$ and $u=1\cdots f$
 \STATE Set i=0, Set $D^0_F=0$
\REPEAT
\STATE i=i+1
  \STATE $G^i_F \leftarrow$ gradient of $J'(\F)+\lambda\Omega(\F)$ wrt.
  $\F$
 \STATE $\beta\leftarrow\frac{\|G^i_F\|^2}{\|G^{i-1}_F\|^2}$ (Fletcher and Reeves)
  \STATE $D^i_F\leftarrow - G^i_F + \beta D^{i-1}_F$
  \STATE $(\F^i,\alpha*)\leftarrow$ Line-Search along $D^i_F$ 
\UNTIL{ Stopping criterion is reached}
\end{algorithmic}
\end{algorithm}

Note that for each computation of $J(F)$ in the line search, the 
optimal $\alpha^*$  is found by solving an SVM. A similar approach, has
been used to solve the Multiple-Kernel problem in
\citep{rakotomamonjy08simplemkl} where the weights of the kernels are
learned by gradient descent and the SVM is solved iteratively. 

At each iteration of the algorithm the gradient of
$J'(\F)+\lambda\Omega(\F)$ has to be computed. With a Gaussian kernel
the gradient of $J'(\cdot)$ wrt. $\F$ is:
\swp{
\begin{align}
  \label{eq:gradgauss}
  \nabla
  J(\F_{u,v})&=&\frac{1}{2\sigma_k}\sum_{i,j}^{n,n}&(\X_{i+1-u,v}-\X_{j+1-u,v})\\
&&&\times(\Xf_{i,v}-\Xf_{j,v})\Kf_{i,j}\y_i\y_j\alpha^*_i\alpha^*_j\nonumber
\end{align}}{
\begin{equation}
  \label{eq:gradgauss}
  \nabla
  J(\F_{u,v})=\frac{1}{2\sigma_k}\sum_{i,j}^{n,n}(\X_{i+1-u,v}-\X_{j+1-u,v})
(\Xf_{i,v}-\Xf_{j,v})\Kf_{i,j}\y_i\y_j\alpha^*_i\alpha^*_j
\end{equation}}
where $\alpha^*$ is the SVM solution for a fixed $\F$. We can
see that the complexity of this gradient is $\mathcal{O}(n^2.f^2)$ but in
practice, SVM have a sparse support vector
representation. So in fact the gradient computation is
$\mathcal{O}(n_s^2f^2)$ with $n_s$ the number of support vector selected.

Due to the non-convexity
of the objective function, it is difficult to provide
an exact evaluation of the solution complexity. 
However, we know that the gradient computation  is
$\mathcal{O}(n_s^2.f^2)$ and that when $J(F)$ is computed in the line
search, a  SVM of size $n$ is
solved and a $\mathcal{O}(n.f.d)$ filtering is applied. Note that a
warm-start trick is used when using iteratively the SVM solver in
order to speed up the method. 

\subsection{Filter regularization}
\label{sec:filt-regul}
In this section we discuss the choice of the filter regularization
term. This choice is important due to the complexity of the KF-SVM
model. Indeed,  learning the FIR filters adds parameters to
the problem and regularization is essential in order to
avoid over-fitting.

The first regularization term that we consider and
use in our KF-SVM framework is the
Frobenius norm:
\begin{equation}
  \label{eq:frobnorm}
  \Omega_2(\F)=\sum_{u,v}^ {f,d}\F_{u,v}^2  
\end{equation}
This regularization term is differentiable and the gradient is easy to
compute. Minimizing this regularization term
  corresponds to minimizing the filter energy. In terms of
classification, the filter matrix can be seen as a kernel parameter
weighting delayed samples. For a given column, such a
sequential weighting is related to a
phase/delay and cut-off frequency of the filter.  Moreover the
Gaussian kernel defined in \Eq \ref{eq:kernelmatrix} shows that the
per column convolution can be seen as a scaling of the channels prior to
kernel computation. 
The intuition of how this regularization term
influences the filter learning is the following.
Suppose we learn our decision function $g(\cdot)$
by minimizing only $J'(.)$, 
the learned filter matrix will maximize
the margin between classes. 
Adding the Frobenius regularizer will force
non-discriminative filter coefficients to vanish thus
yielding to reduced impact on the kernel of some delayed samples.

Using this regularizer, all filter coefficients are treated 
independently, and  even if it
tends to down-weight some  non-relevant channels,
filter coefficients are not sparse.
If we want to perform a channel selection while learning the filter $\F$,
we have to force some columns of $\F$ to be zero. For that, we can use
a $\ell_1-\ell_2$ mixed-norm as a regularizer:
\begin{equation}
  \label{eq:mixednorm}
  \Omega_{1-2}(\F)=\sum_{v}^ {d}\left( \sum_u^f\F_{u,v}^2  \right)^{\frac{1}{2}}=\sum_{v}^ {d}h\left( ||\F_{.,v}||^2 \right)
\end{equation}
with $h(x)=x^{\frac{1}{2}}$ the square root function. Such a mixed-norm 
acts as a $\ell_2$ norm on each
single channel filter while the $\ell_1$ norm on each channel
filter energy will tend to vanish all coefficients
related to a channel.
As this regularization term is not differentiable, 
the solver proposed in Algorithm~\ref{algorithm1} can not be used. 
We address the
problem through a Majorization-Minimization
algorithm \citep{hunter2004tutorial} that enables us to take advantage of
the KF-SVM solver proposed above. 
The idea here is to iteratively replace $h(\cdot)$ by a majorization
and to minimize the resulting objective function.
Since $h(\cdot)$ is concave in its positive orthant,
we consider the following linear majorization of
$h(\cdot)$ at a given point $x_0>0$ : 
$$ \forall x>0,\quad\quad h(x) \leq x_0^\frac{1}{2} + \frac{1}{2} x_0^{-\frac{1}{2}}(x-x_0) $$
The main
advantage of a linear majorization is that we can re-use KF-SVM
algorithm.
Indeed, at iteration $k+1$, for $\F^{(k)}$ the solution at iteration $k$,
applying this linear majorization of
$h(\|\F_{\cdot,v}\|)$, around a $\|\F_{\cdot,v}^{(k)}\|$ yields to a
Majorization-Minimization algorithm for sparse filter learning
 which consists in\swp{}{ iteratively} solving:
 \begin{align}
   \label{eq:mmalgo}
  &\min_{\F^{(k+1)}} J'(\F)+\lambda\Omega_\w(\F)&\\
&\text{with } \Omega_{\w}(\F)=\sum_{v}^ {d}\w_j
\sum_u^f\F_{u,v}^2\text{ and }\w_v=\frac{1}{\|\F_{.,v}^{(k)}\|}&\nonumber
 \end{align}
$\Omega_{\w}$ is a weighted Frobenius norm,
this regularization term is differentiable and the KF-SVM solver can be
used. We call this method Sparse
KF-SVM (SKF-SVM) and \swp{the solver can be seen in Algorithm~\ref{algorithm2}.
We}{we} use here similar stopping criteria as in Algorithm~\ref{algorithm1}.
\swp{
 \begin{algorithm}[t]
 \caption{ SKF-SVM solver \label{algorithm2}}
 \begin{algorithmic}
 \STATE Set $\F_{u,v}=1/f$ for $v=1\cdots d$ and $u=1\cdots f$
 \STATE Set $\w_k=1$ for $k=1\cdots d$
\REPEAT
  \STATE $(\F,\alpha)\leftarrow$ Solve KF-SVM with $\w$ column weights
  \STATE $\w_k\leftarrow\frac{1}{||\F_{.,k}||}$ for $k=1\cdots d$
\UNTIL{ Stopping criterion is reached}
\end{algorithmic}
\end{algorithm}
}{}

\subsection{Online and Viterbi decoding}
\label{sec:viterbi-decoding}

In this section, we discuss the decoding complexity of our
method in two cases: when using only the sample classification
score for decision and when using an offline Viterbi decoding of the
complete sequence.

First we discuss the online decoding complexity. The
multi-class case is handled by One-Against-One strategy. So in order
to decide the label of a given sample, the score for each class has to
be computed with the decision function \eqref{eq:5} that is
$\mathcal{O}(n_s)$ with $n_s$ the number of support vectors. Finally
the decoding of a sequence of size $n$ is  $\mathcal{O}(n_s.c.n)$ with
$c$ the number of classes.


The offline Viterbi decoding relies on the work of
\citet{ganapathiraju2004applications} who proposed to transform the output of
SVM classifiers into probabilities with a sigmoid function
\citep{lin2007note}.
 The estimated probability for class $k$ is:
\begin{equation}
  \label{eq:platt}
  P(y==k|x)=\frac{1}{1+\exp{\left(A.g_k(x)+B\right)}}
\end{equation}
where $g_k$ is the One-Against-All decision function for class $k$ and $x$ the
observed sample. $A$ and $B$ coefficients are learned by
maximizing the log-likelihood
on a validation set. 
The inter-class transition matrix $M$ is estimated on the learning
set. 
Finally the Viterbi
algorithm is used to obtain the maximum likelihood sequence. The
complexity for a sequence of size $n$ is then $\mathcal{O}(n_s.c.n)$ to obtain the
pseudo-probabilities and $\mathcal{O}(n.c^2)$ to decode the sequence.

\subsection{Related works}
\label{sec:disc-relat-works}

To the best of our knowledge, there has been few
works dealing with the joint learning of a temporal filter and a decision function. 
The first one addressing such a problem is our 
work \citep{flamaryicassp210} that solves
the problem for linear decision functions.
Here, we have extended this approach to the non-linear case and
we have also investigated the utility of different regularizers
on the filter coefficients. Notably, we have introduced
regularizers that help in performing channel selection.

Works on Common Sparse Spatio-Spectral Patterns \cite{dornhege2006optimizing} are probably those 
that are the most similar to ours. Indeed,
they  want to learn a linear combination of channels
and samples that optimize a separability criterion.
But the criterion optimized by the two
algorithms are different: CSSSP aims at maximizing the variance
of the samples for the positive class while minimizing the variance
for the negative class, whereas KF-SVM aims at maximizing the margin between
classes in the feature space. 
Furthermore, CSSSP is a feature extraction algorithm
that is independent to the used classifier whereas in our
case, we learn a filter that is tailored to the
(non-linear) classification algorithm criterion. 
Furthermore, the filter used in KF-SVM is not restricted
to signal time samples but can also be applied
to complex sequential features extracted from the
signal (\emph{e.g} PSD). An application of
this latter statement is provided in the experimental
section.

KF-SVM can also be seen as a kernel learning
method.
~Indeed the filter coefficients can be interpreted as kernel parameters
despite the fact that samples are non-iid.
Learning such a kernel
parameters is now a common approach
introduced by \cite{chapelle_tuning}. 
While \citeauthor{chapelle_tuning} minimize a bound on
generalization error by gradient descent, in our
case we simply minimize the SVM objective 
function and the influence on the parameters
differ. 
More precisely, if we focus on the colums of $\F$ we see that the
coefficients of these columns act as a scaling of the
channels. For a filter of size $1$, our approach
would correspond to adaptive scaling as
proposed by
\citet{grandvalet_adaptive}. In their work, 
they jointly learn the
classifier and the Gaussian kernel parameter $\sigma_k$ with a
sparsity constraint on the dimensions
of $\sigma_k$ leading to automated feature selection. KF-SVM can
thus be seen as a
generalization of their approach which takes into account samples
sequentiality.

\section{Numerical experiments}
\label{sec:results}

\swp{
Numerical experiments are presented for a toy example and real life
BCI Dataset from \emph{BCI Competition
  III}~\cite{bcicometitioniii}.}{}

\subsection{Toy Example}
\label{sec:toy-example}

In this section we present the toy example used for numerical
experiments. Then we discuss the performances and the parameter
sensitivity of our method.

\label{sec:toy-example-defin}

 \begin{figure}[t]
   \centering
   \swp{\includegraphics[width=\linewidth]{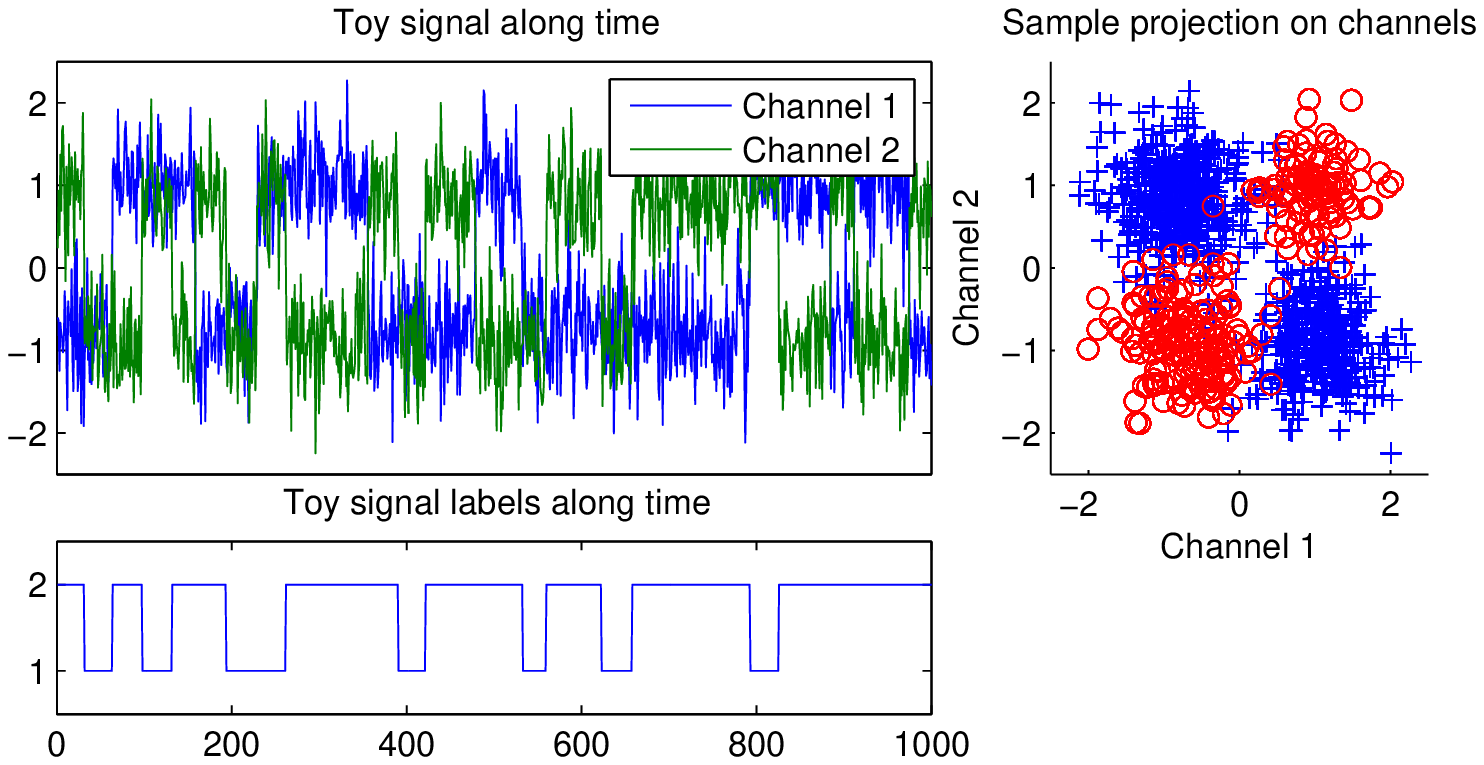}}{
   \includegraphics[width=.6\linewidth]{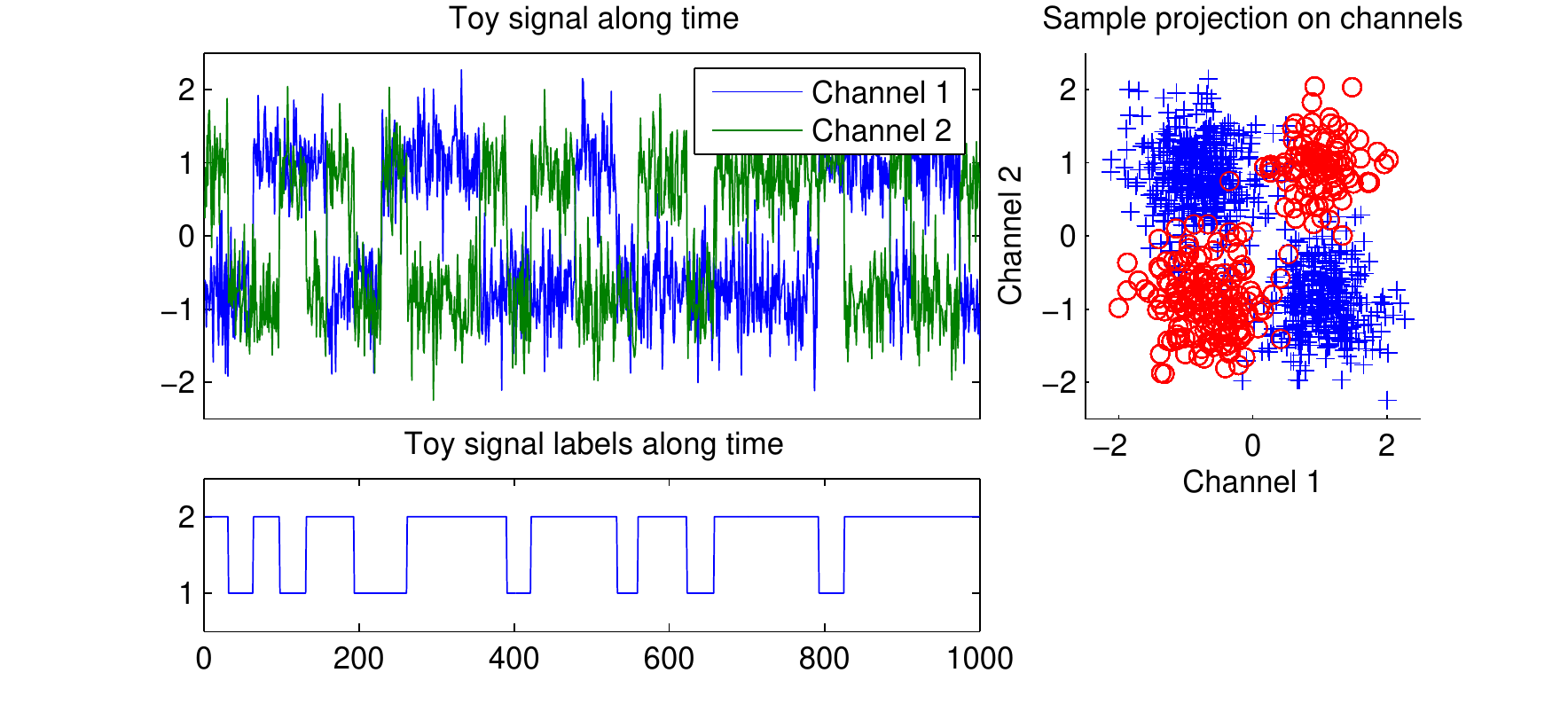}}
   \caption{Toy example for $\sigma_n=0.5,lag=0$. The plots on the
     left show the evolution of both channels and labels along time; 
the right plot shows the non-linear problem by projecting the
   samples on the channels.}
   \label{fig:toyexample}
 \end{figure}
We use a toy example that consists of a 2D non-linear problem
which can be seen on  Figure~\ref{fig:toyexample}.
Each class contains 2
modes, $(-1,-1)$ and $(1,1)$ for class 1 and $(-1,1)$ and
$(1,-1)$ for class 2, and their value is corrupted by a Gaussian noise
of deviation $\sigma_n$. Moreover, the length of the regions with constant label follows a
uniform distribution between $[30,40]$~samples. A time-lag drawn from
a uniform distribution between $[-lag,lag]$ is applied to the channels
leading to mislabeled samples in the learning and test set.

\begin{figure}[t]
  \centering
 \subfigure[Without filtering (err=0.404)]{
\swp{\includegraphics[width=.9\linewidth]{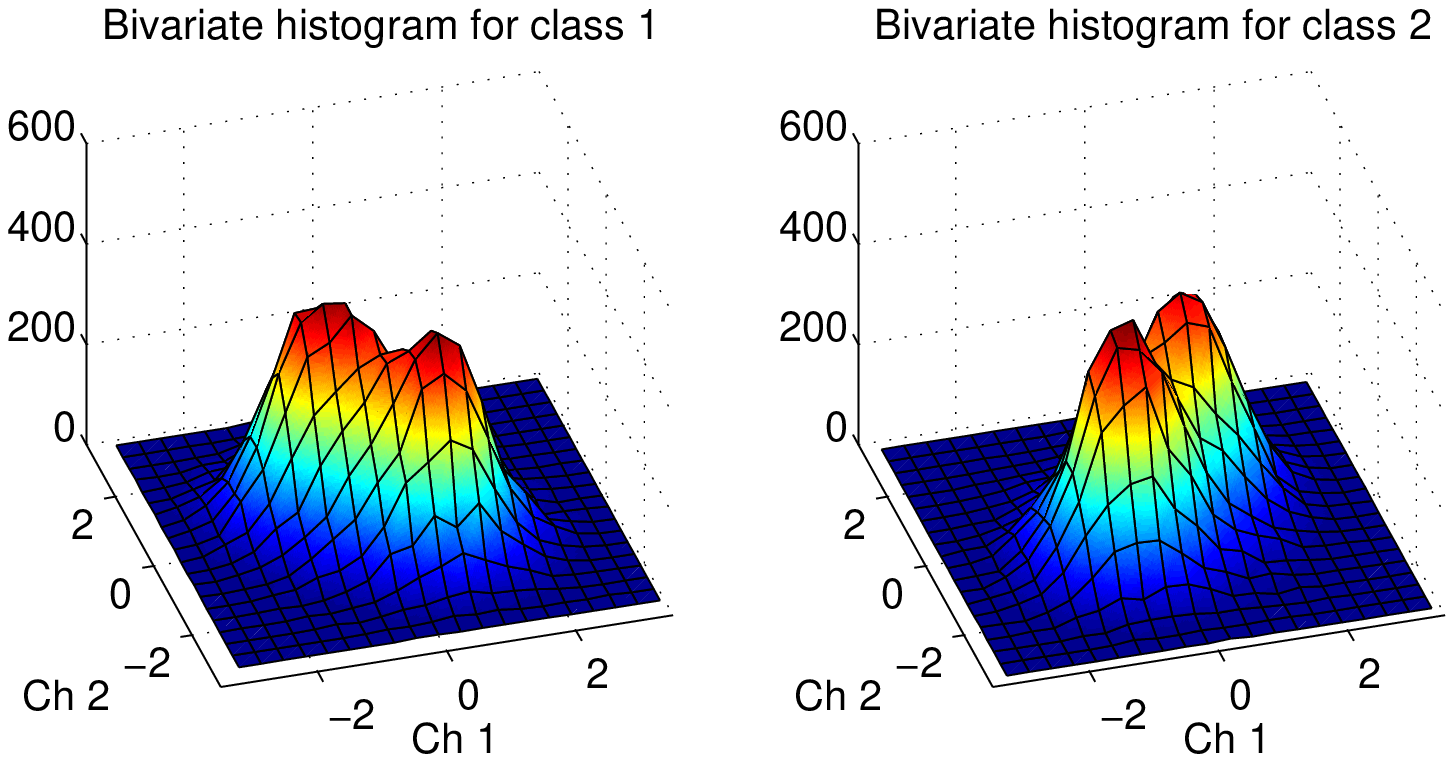}}{
\includegraphics[width=.45\linewidth]{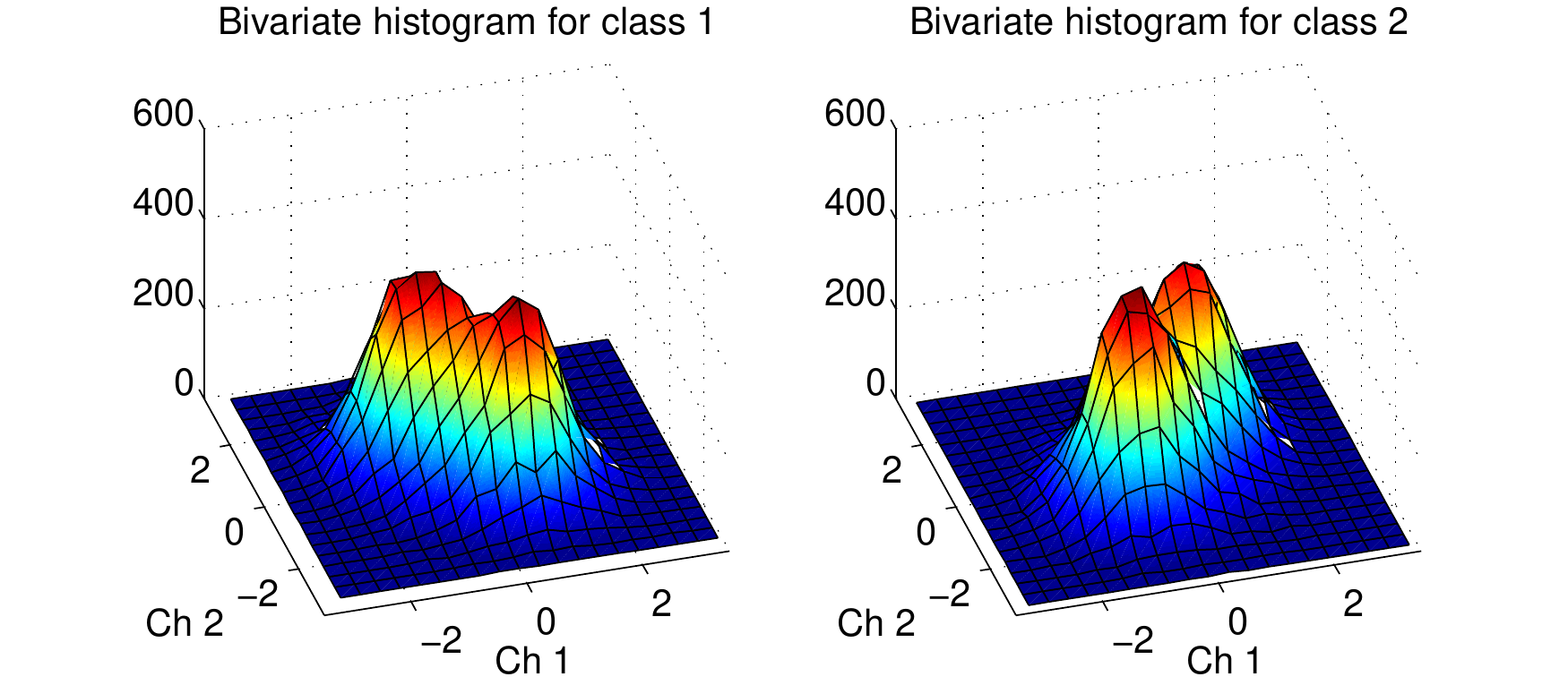}}\label{fig:non_filtered}}
\subfigure[With KF-SVM filtering (err=0.044)]{
\swp{\includegraphics[width=.9\linewidth]{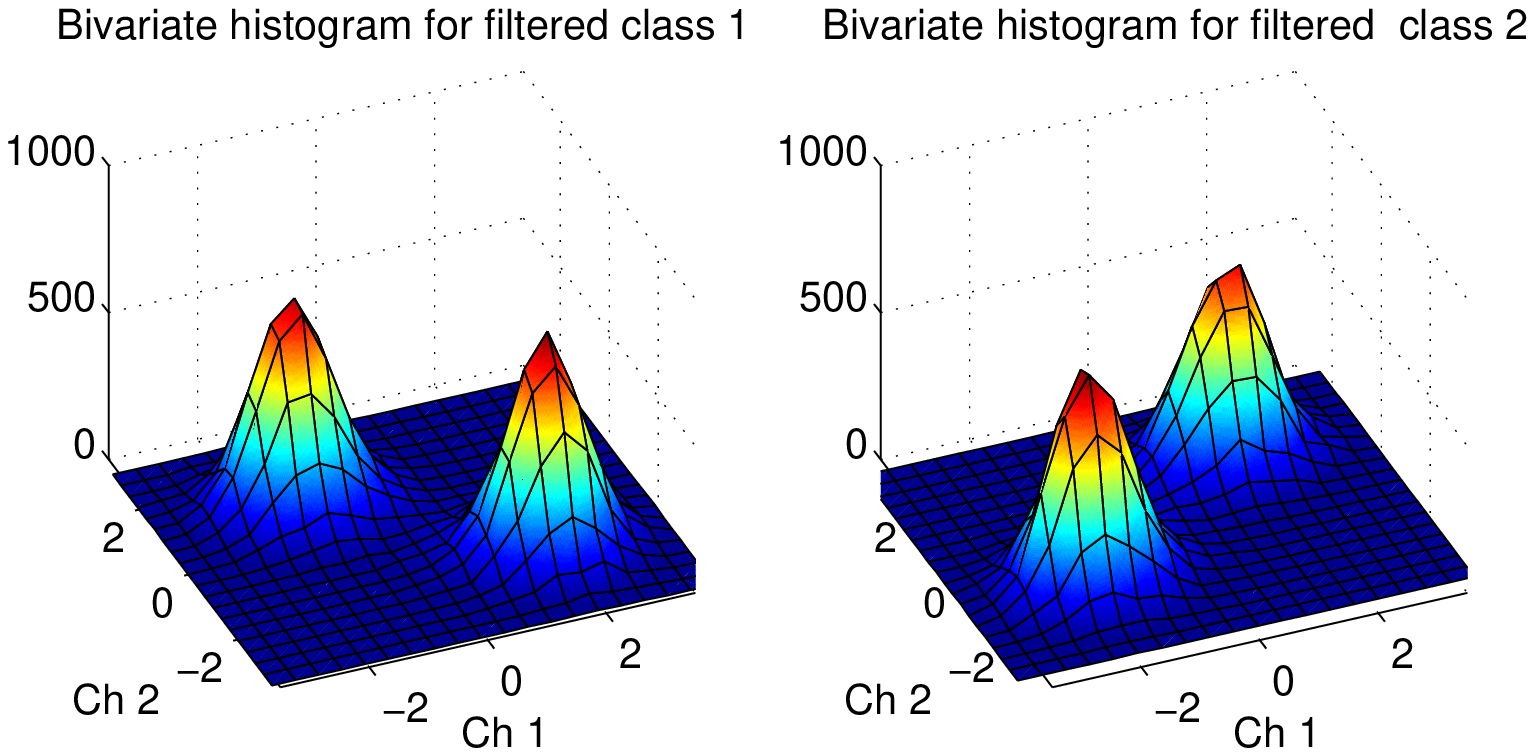}}{
\includegraphics[width=.45\linewidth]{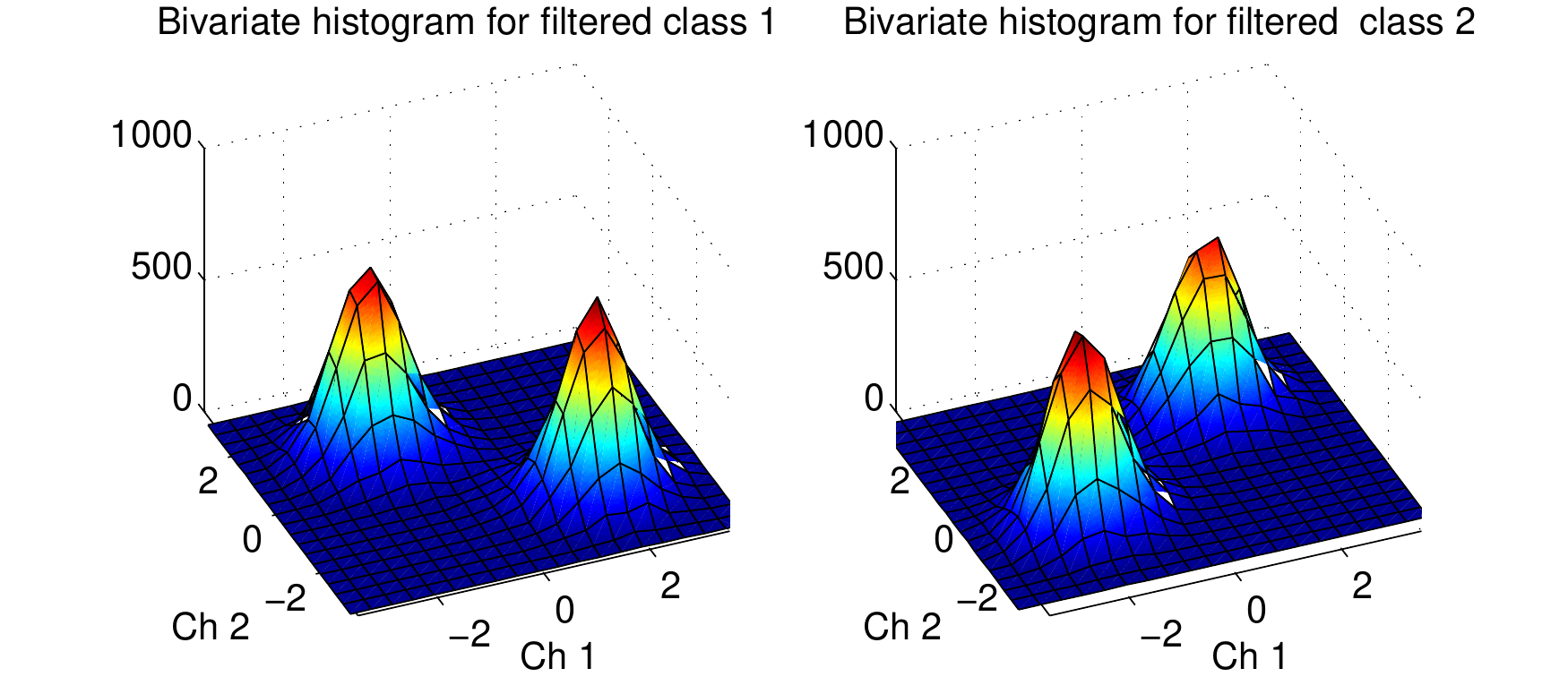} } \label{fig:filtered}}
  \caption{Bivariate histograms for a non filtered ($\sigma_n=1,lag=5$) and KF-SVM filtered
    signal (left for class 1 and right for class 2)}
  \label{fig:bivariate_hist}
\end{figure}

We illustrate the behavior of the large margin filtering on a simple
example ($\sigma_n=1,lag=5$). The bivariate histogram of the
projection of the samples on the channels can be seen
on Figure~\ref{fig:bivariate_hist}. 
We can see on Figure~\ref{fig:non_filtered} that due to the noise and
time-lag there is an important overlap between the bivariate
histograms of both classes, but when the large margin filter is
applied, the classes are better
separated (Figure~\ref{fig:filtered})  and the overlap is reduced leading to better 
classification rate (4\% error vs 40\%).

\swp{
\subsubsection{Classification performances}
\label{sec:class-perf}}{}

\swp{
 \begin{figure*}[ht]
    \centering
\includegraphics[width=.65\columnwidth]{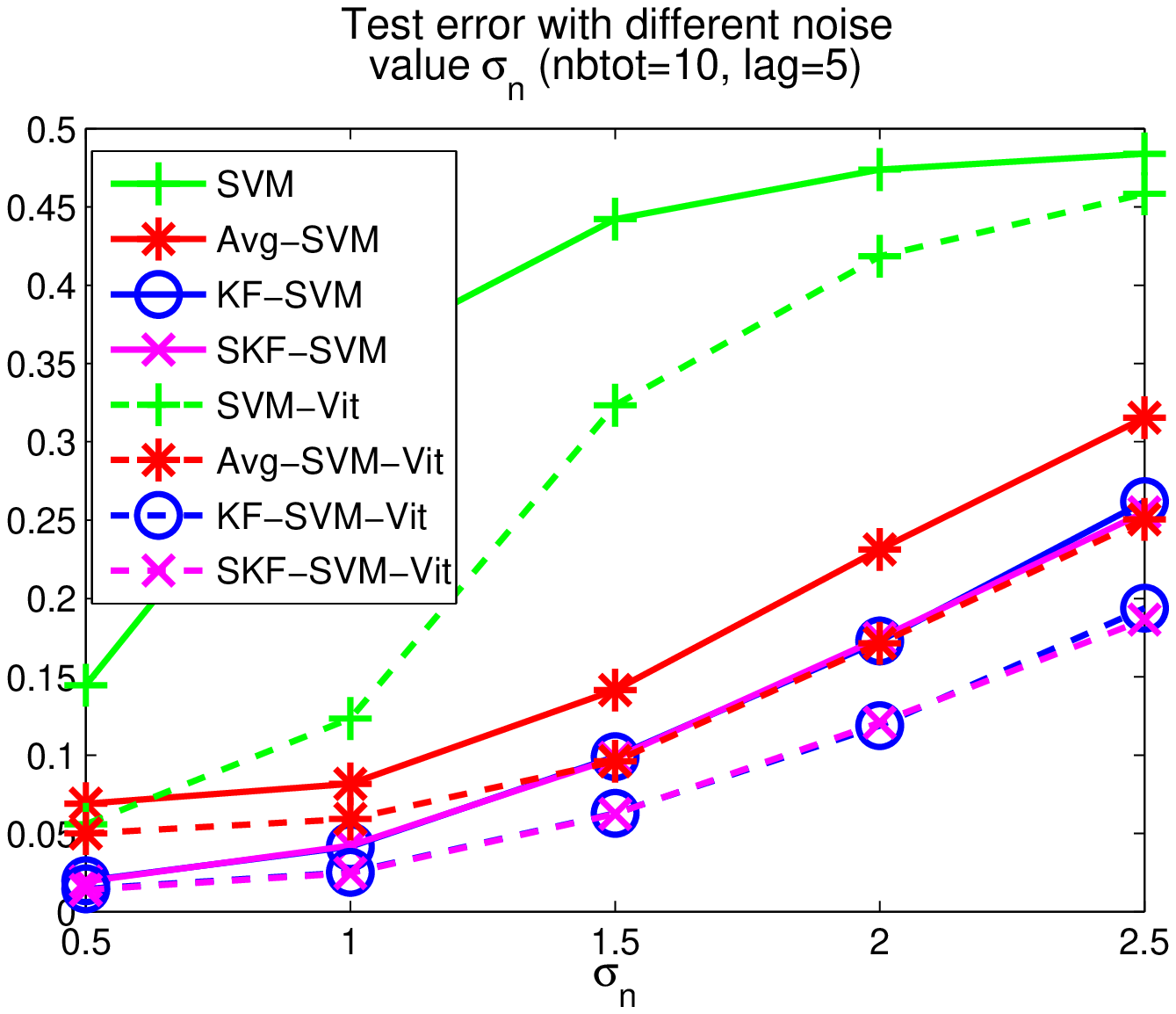}
   \includegraphics[width=.65\columnwidth]{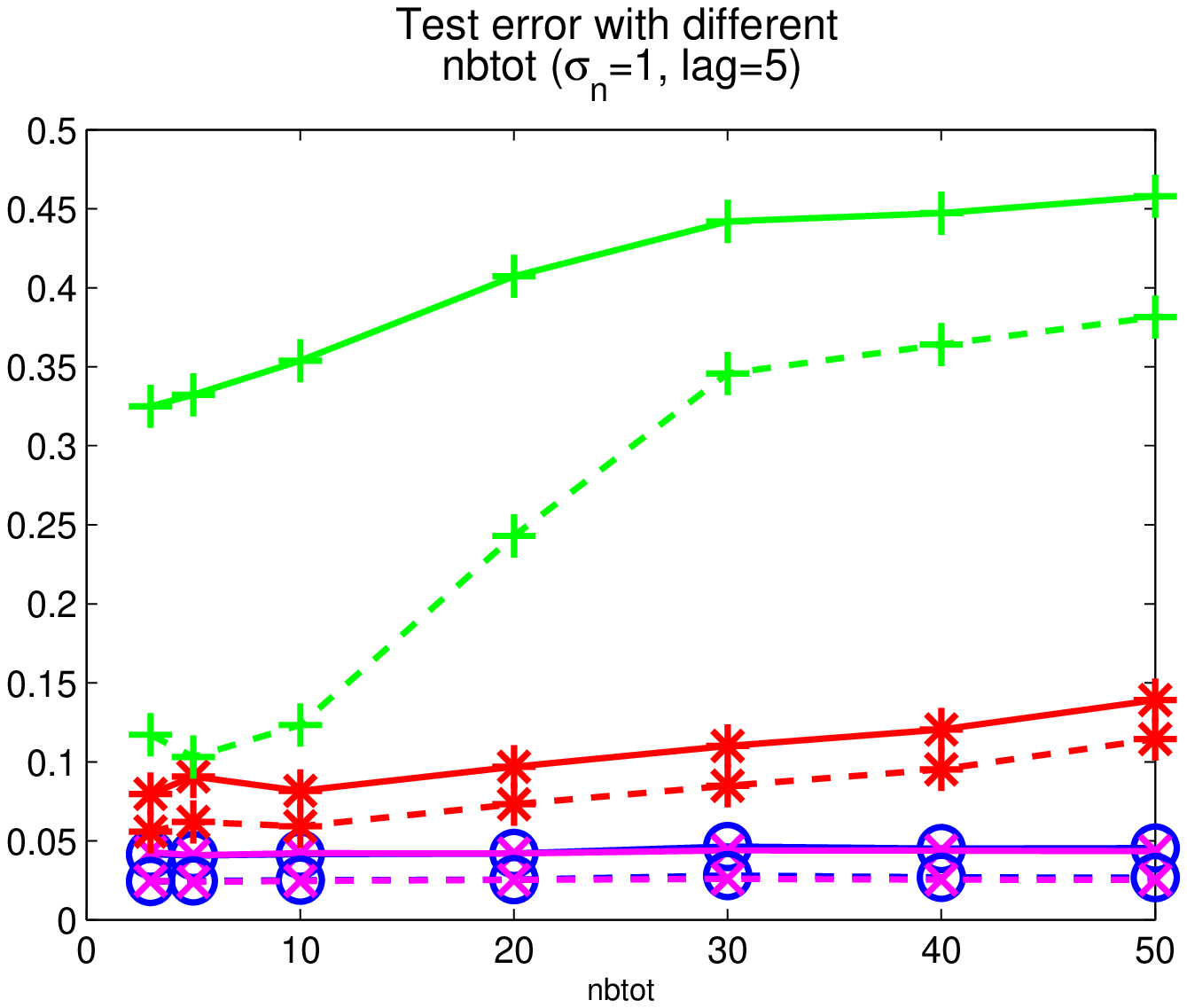} 
\includegraphics[width=.65\columnwidth]{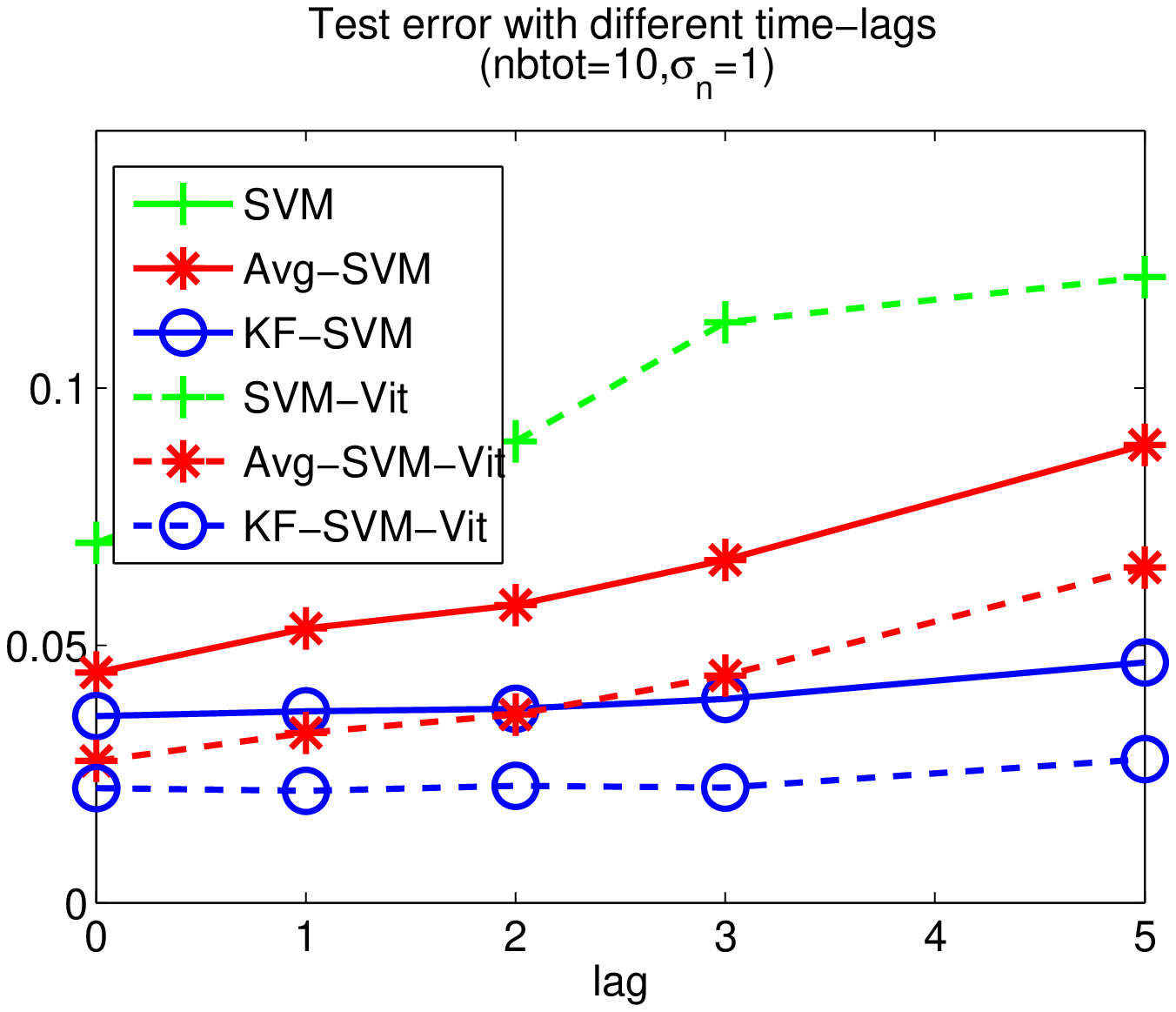}
   \caption{Test error for the toy example with different noise value
     (left), problem size (middle) and time-lag (right)\quad\quad (plain lines :
     sample classification , dashed lines :
     Viterbi decoding)}
   \label{fig:res_toy} 
\end{figure*}
}{
 \begin{figure*}[ht]
    \centering

   \subfigure[varying
   noise value $\sigma_k$]{\swp{\includegraphics[width=.65\columnwidth]{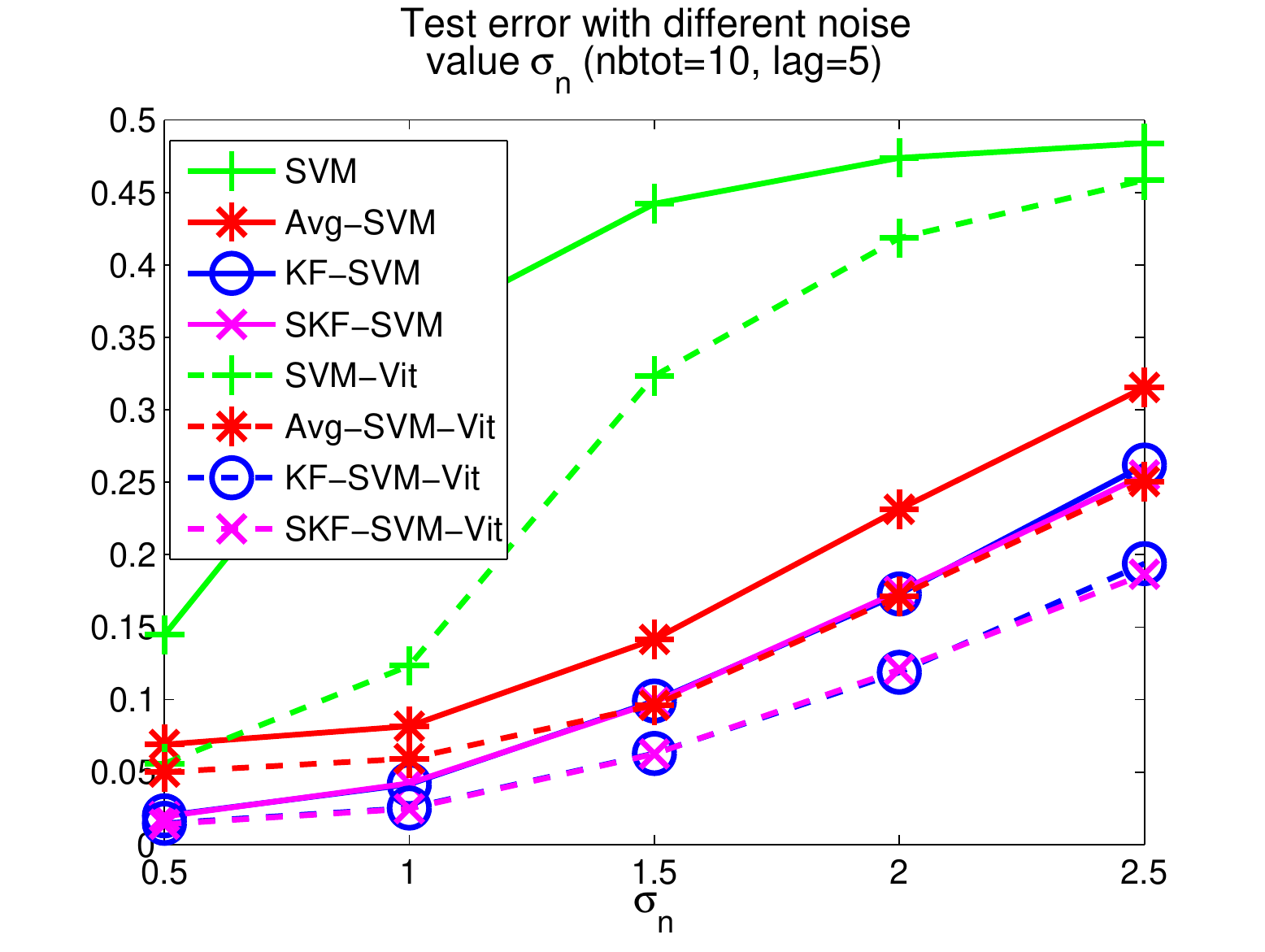}}{
\includegraphics[width=.4\columnwidth]{imgs/toy_noise}}\label{fig:toy_noise}}
\swp{\hspace{2cm}}{}
   \subfigure[varying
   size $nbtot$ of the problem]{\swp{\includegraphics[width=.65\columnwidth]{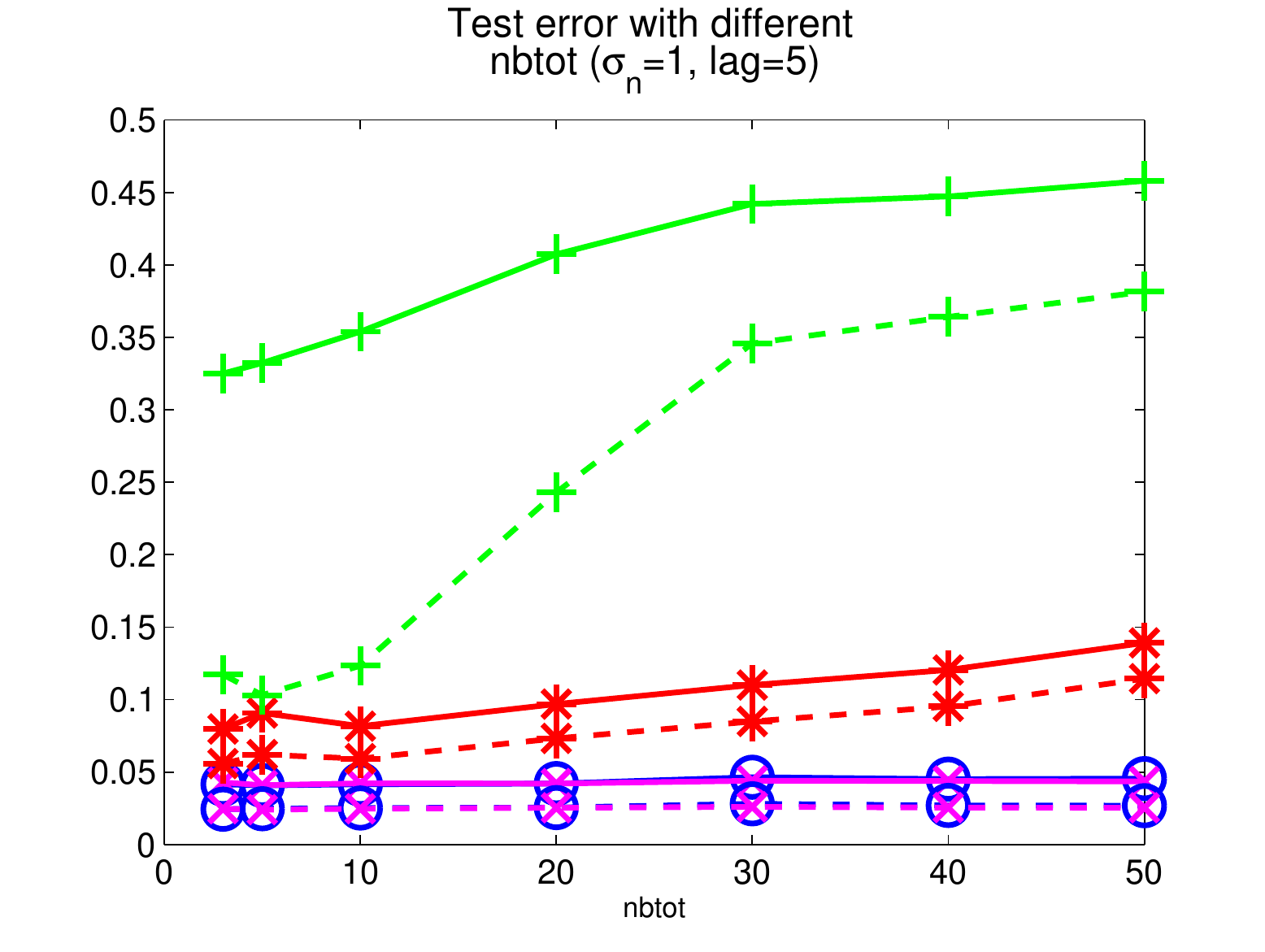}}{
\includegraphics[width=.4\columnwidth]{imgs/toy_size}}\label{fig:toy_size}}
   \caption{Test error for different problem size and noise for the toy example (plain lines :
     sample classification , dashed lines :
     Viterbi decoding)}
   \label{fig:res_toy}
 \end{figure*}}

SVM, Avg-SVM (signal filtered by average filter), KF-SVM and SKF-SVM
are compared with and without Viterbi decoding. In order to test high
dimensional problems, some channels containing only gaussian noise are added to the 2
discriminative ones leading to a toy signal of $nbtot$ channels.
The size of the signal is of $1000$~samples for the learning
and the validation sets and of $10000$~samples for the test set.
In order to compare fairly with Avg-SVM, we selected $f=11$  and
$n_0=6$ corresponding to a good average filtering centered on the
current sample. 
The regularization parameters are selected by a
validation method. All the processes are run ten times, the test error is then the average
over the runs.

We can see in Figure~\ref{fig:res_toy} the test error for different
noise value $\sigma_n$\swp{,}{and} problem size $nbtot$\swp{ and time-lag $lag$}{}.
Both proposed methods outperform SVM and Avg-SVM with a Wilcoxon
signed-rank test p-value$<0.01$. Note that results obtained with KF-SVM
without Viterbi decoding are even better than those observed with SVM
and Viterbi decoding. This is probably because as we said previously,
HMM can not adapt to time-lags because the learned
density estimation are biased.
Surprisingly, the use of the sparse
regularization does not statistically improve the results despite the
intrinsic sparsity of the problem. This comes from the fact that the learned filters
of both methods are sparse due to a numerical precision thresholding
for KF-SVM with Frobenius regularizer. Indeed the $\lambda$
coefficient selected by the validation is large, leading to a
shrinkage of the non-discriminative channels.

\swp{
\subsubsection{Parameters sensitivity}
\label{sec:param-sens}}{}

We discuss the importance of the choice of our model
parameters. In fact KF-SVM has 4 important parameters that have
to be tuned: $\sigma_k$, $C$, $\lambda$ and $f$. Those parameters have to be tuned in
order to fit the problem at hand. Note that $\sigma_k$ and $C$ are
parameters linked to the SVM approach and that the remaining ones are
due to the filtering approach. In the results presented
below, a validation has been done to select $\lambda$ and $C$.

\swp{
\begin{figure*}[ht]
   \centering
\includegraphics[width=.65\columnwidth]{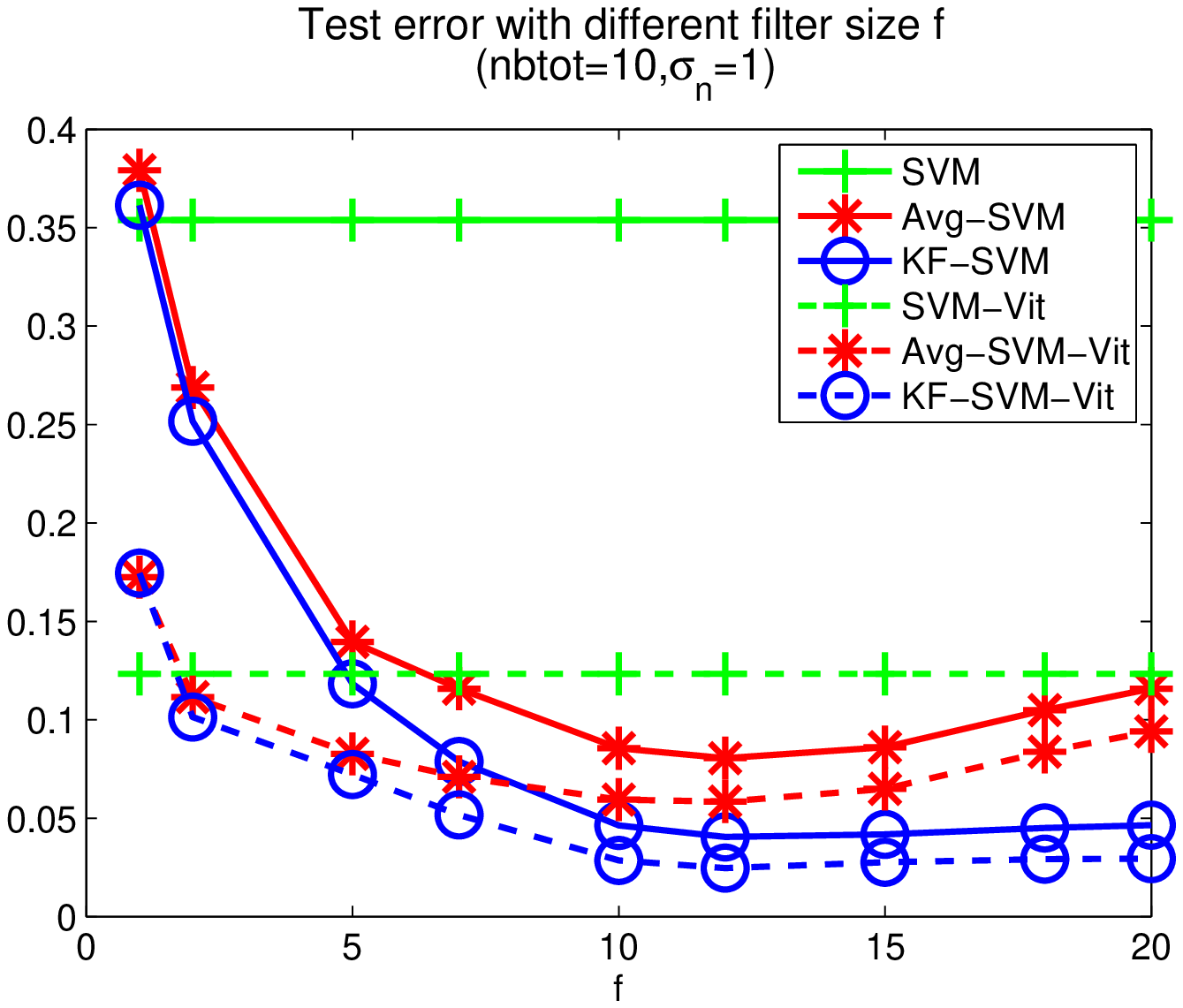}\hspace{.2\columnwidth}
\includegraphics[width=.65\columnwidth]{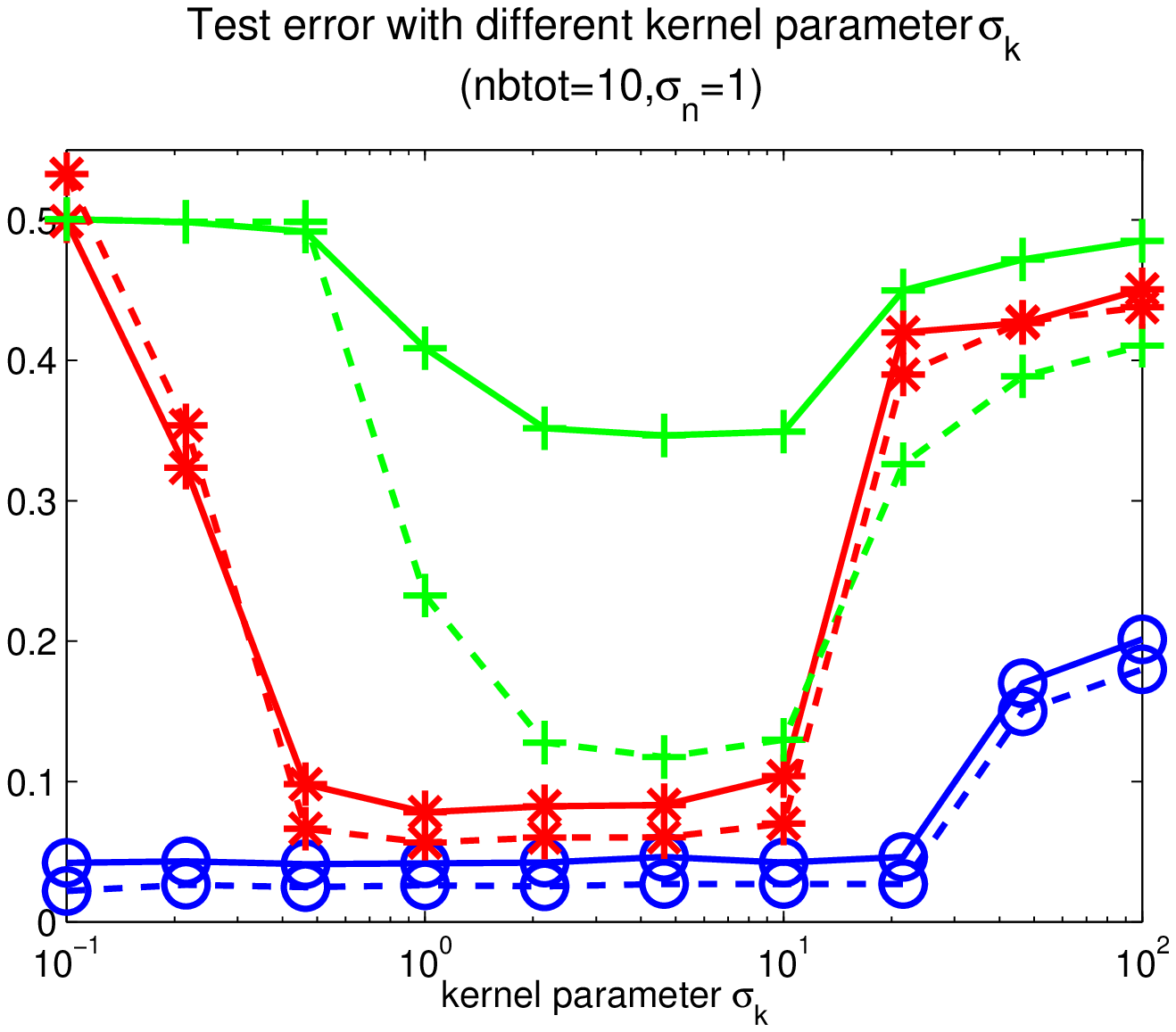}
   \caption{Test error for the toy example with different filter size
     $f$ (on the left) and kernel parameter $\sigma_k$ (on the right) \quad(plain lines :
     sample classification , dashed lines :
     Viterbi decoding)}
   \label{fig:res_toy_params}
 \end{figure*}}{
\begin{figure*}[ht]
   \centering
   \subfigure[varying
   f]{\swp{\includegraphics[width=.7\columnwidth]{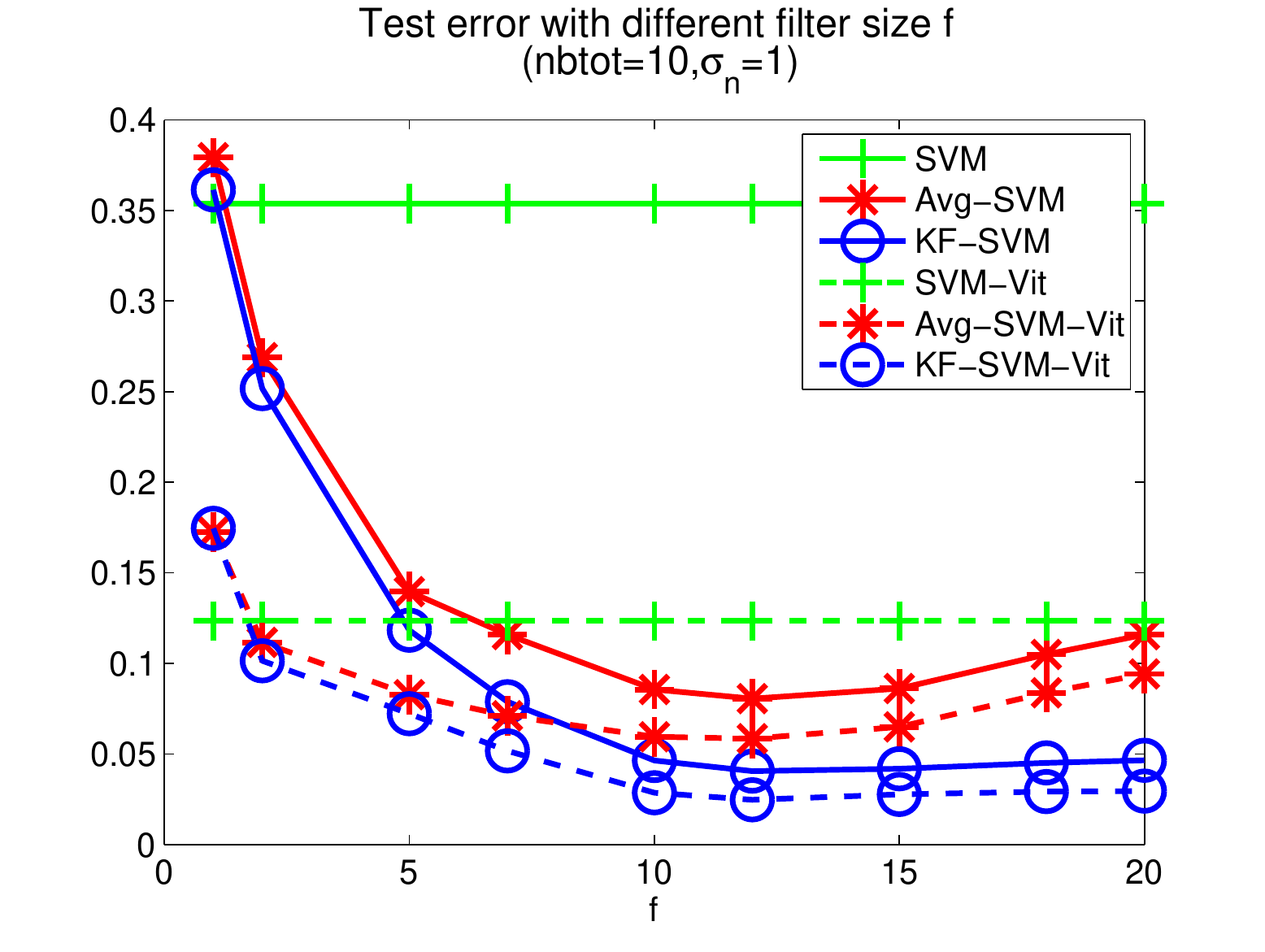}}{
\includegraphics[width=.4\columnwidth]{imgs/toy_f}}\label{fig:toy_f}}
\swp{\hspace{2cm}}{}
   \subfigure[varying
   sigma]{\swp{\includegraphics[width=.7\columnwidth]{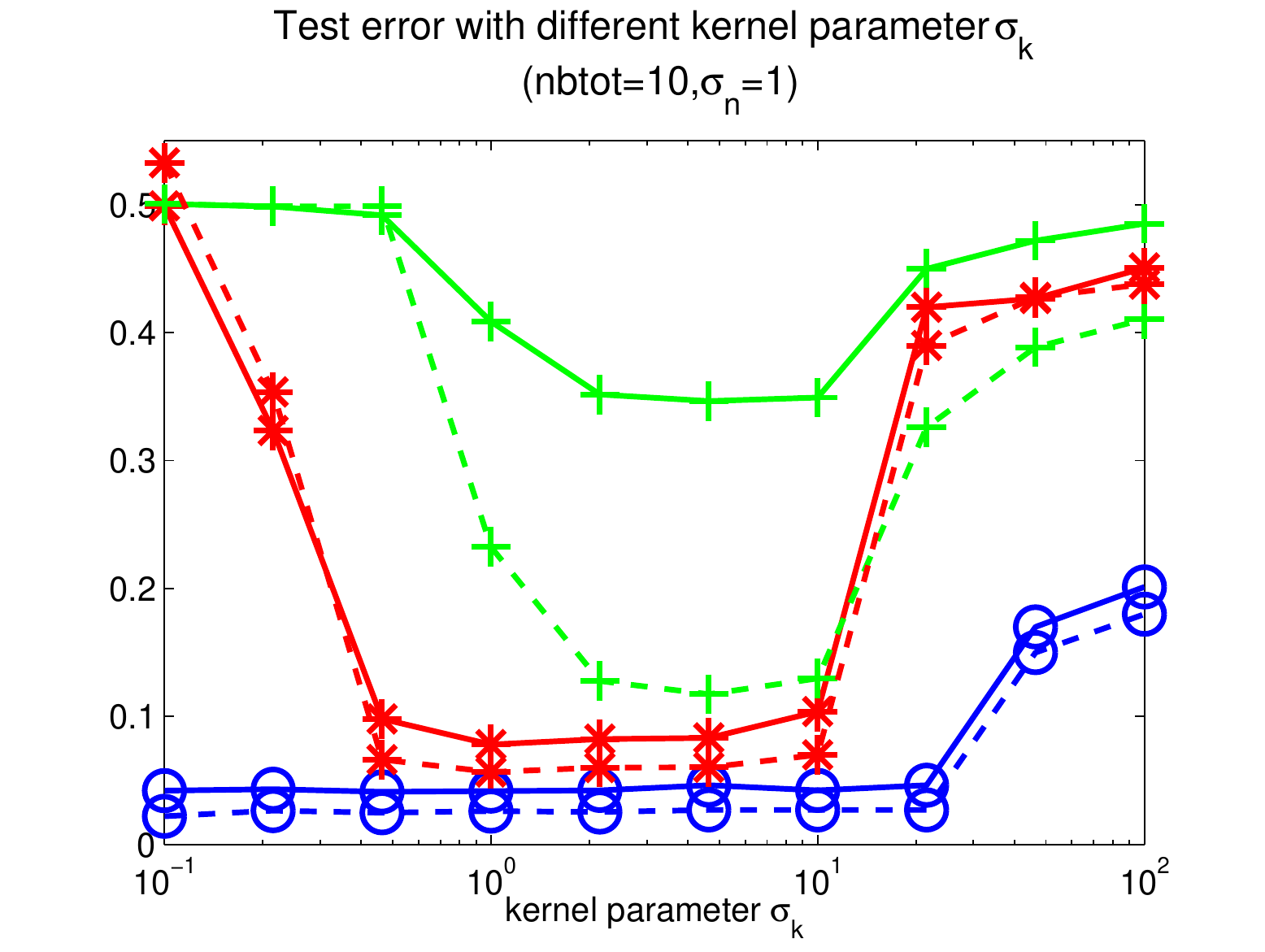}}{
\includegraphics[width=.4\columnwidth]{imgs/toy_sigma}}\label{fig:toy_sigma}}
   \caption{Test error for different parameters on the toy example (plain lines :
     sample classification , dashed lines :
     Viterbi decoding)}
   \label{fig:res_toy_params}
 \end{figure*}

}

We can see on the left of Figure \ref{fig:res_toy_params} the
performances of the different models for a varying $f$. Note that $f$
has a big impact on the performances when using Avg-SVM. On the
contrary,  KF-SVM shows good performances for a
sufficiently long filter, 
due to the learning of the filtering. Our
approach is then far less sensitive to the size of the filter than
Avg-SVM. 
Finally we discuss the sensitivity to the kernel parameter $\sigma_k$.
Test errors for different values of this parameters are shown on Figure
\ref{fig:res_toy_params} (right). It is interesting to note that KF-SVM is far less
sensitive to this parameter than the other methods. Simply because the
learning of the filtering corresponds to an automated scaling of the
channels which means that if the $\sigma_k$ is small enough the
scaling of the channels will be done automatically.
\swp{
}{}
In conclusion to these results, we can say that despite the fact that
our method has more parameters to tune than a simple SVM approach, it
is far less sensitive to two of these parameters than SVM.

\subsection{BCI Dataset}
\label{sec:bci-dataset}

We test our method on the BCI Dataset from \emph{BCI Competition
  III}~\citep{bcicometitioniii}. The
problem is to obtain a sequence of labels out of brain activity
signals for 3 human subjects. The data consists in 96~channels
containing PSD features (3~training sessions, 1~test session,
$n\approx{3000}$ per session) and the
problem has 3~labels (left~arm, right~arm or a word). 

\label{sec:performances}

For computational reasons, we decided to decimate the signal by 5,
doing an averaging on the samples. We focus on online
sample labeling ($n_0=0$) and we test KF-SVM for filter length $f$
corresponding to those used in \citep{flamaryicassp210}.
The regularization
parameters are tuned using a grid search validation method on the third
training set. Our method is compared to the best BCI competition
results  and to the SVM without filtering. Test error for
different filter size $f$ can be seen on Table \ref{tab:bcidataset}. 
\begin{table}[t]
 \swp{ \caption{Test Error for BCI Dataset }
  \label{tab:bcidataset}
\vskip 0.15in}{}

  \centering
\begin{small}
\begin{sc}
\begin{tabular}{|l|c|c|c|c|} \hline
Method & Sub 1 & Sub 2 & Sub 3 & Avg\\
\hline
BCI Comp.   & 0.2040 & 0.2969 & \textbf{0.4398} & 0.3135 \\ \hline
SVM         & 0.2368 & 0.4207 & 0.5265 & 0.3947 \\ \hline
KF-SVM  &        &        &        &        \\
$f=2$ &    0.2140  &  0.3732  &  0.4978  &  0.3617\\
$f=5$ &    0.1840  &  0.3444  &  0.4677  &  0.3320\\
$f=10$&    \textbf{0.1598}  & \textbf{ 0.2450}  &  0.4562  &  \textbf{0.2870}\\ \hline
\end{tabular}
\end{sc}
\end{small}

 \swp{\vskip -0.2in}{ \caption{Test Error for BCI Dataset }
  \label{tab:bcidataset}}
\end{table}
We can see that we improve the BCI Competition results by using
longer filtering. We obtain similar results than those reported in
\cite{flamaryicassp210} but slightly worst. This probably comes
from the fact that the features used in this Dataset are PSD and are
known to work well in the linear case. But we still obtain competitive
results which is promising in the case of non-linear features.

\label{sec:filter-visualisation}

\section{Conclusions}
\label{sec:conclusion}

We have proposed a framework for learning large-margin filtering for non-linear
multi-channel sample labeling. Depending on the regularization term
used, we can do either an adaptive scaling of the channels or a
channels selection. We proposed a conjugate
gradient algorithm to solve the
minimization problem and empirical results showed that despite the
non-convexity of the objective function our approach performs better
than classical SVM methods.
We tested our approach on a non-linear toy example and on a real life
BCI dataset and we showed that
sample classification rate and precision after Viterbi decoding can be
drastically improved. Furthermore we studied the
sensitivity of our method to the regularization parameters.

In future work, we will study the use of prior information on the
classification task. For instance when we know that the noise is in
high frequencies then we could force the filtering to be a low-pass
filter. In addition, we will address the problem of computational learning
complexity as our approach is not suitable to large-scale problems.







\small
\bibliography{biblioRF,biblioAR}

\end{document}